\documentclass[10pt,twocolumn,letterpaper]{article}

\usepackage{iccv}
\usepackage{times}
\usepackage{epsfig}
\usepackage{graphicx}
\usepackage{amsmath}
\usepackage{amssymb}
\usepackage{booktabs}
\usepackage{bbm}
\usepackage{stfloats}
\usepackage{multicol}
\usepackage{multirow}
\usepackage{array, caption, threeparttable}
\usepackage{ulem}
\usepackage{amsbsy}

\usepackage[T1]{fontenc}
\usepackage[utf8]{inputenc}
\usepackage{authblk}
\usepackage{newunicodechar}
\newunicodechar{ﬁ}{fi}
\newunicodechar{ﬂ}{fl}

\usepackage[breaklinks=true,bookmarks=false,colorlinks,urlcolor=black]{hyperref}

\iccvfinalcopy % *** Uncomment this line for the final submission

\def\iccvPaperID{7410} % *** Enter the ICCV Paper ID here

\ificcvfinal\fi

\begin{document}

%%%%%%%%% TITLE
\title{SOTR: Segmenting Objects with Transformers}

\newcommand*\samethanks[1][\value{footnote}]{\footnotemark[#1]}
\author[1,3]{Ruohao Guo\thanks{Equal contribution.}}
\author[2]{Dantong Niu\samethanks}
\author[1,3]{Liao Qu}
\author[1,3,4]{Zhenbo Li\thanks{Corresponding author. E-mail:
\href{mailto:lizb@cau.edu.cn}{lizb@cau.edu.cn}}}
\affil[1]{College of Information and Electrical Engineering, China Agricultural University}
\affil[2]{EECS department, University of California, Berkeley}
\affil[3]{Key Laboratory of Agricultural Information Acquisition Technology, Ministry of Agriculture}
\affil[4]{National Innovation Center for Digital Fishery, China Agricultural University}

% \\
% % China Agricultural University\\
% % {\tt\small ruohguo@foxmail.com}
% \and
% Dantong Niu\\
% % University of California, Berkeley\\
% % {\tt\small niudantong.88@gmail.com}

% \and
% Liao Qu\\
% % China Agricultural University\\
% % {\tt\small liaoqu@cau.edu.cn}

% \and
% Zhenbo Li\\
% China Agricultural University\\
% {\tt\small lizb@cau.edu.cn}
% }

\maketitle
\ificcvfinal\thispagestyle{empty}\fi

%%%%%%%%% ABSTRACT
\begin{abstract}
% The ABSTRACT is to be in fully-justified italicized text, at the top of the left-hand column, below the author and affiliation information. Use the word ``Abstract'' as the title, in 12-point Times, boldface type, centered relative to the column, initially capitalized. The abstract is to be in 10-point, single-spaced type. Leave two blank lines after the Abstract, then begin the main text. Look at previous ICCV abstracts to get a feel for style and length.
Most recent transformer-based models show impressive performance on vision tasks, even better than Convolution Neural Networks (CNN). In this work, we present a novel, flexible, and effective transformer-based model for high-quality instance segmentation. The proposed method, Segmenting Objects with TRansformers (SOTR), simplifies the segmentation pipeline, building on an alternative CNN backbone appended with two parallel subtasks: (1) predicting per-instance category via transformer and (2) dynamically generating segmentation mask with the multi-level upsampling module. SOTR can effectively extract lower-level feature representations and capture long-range context dependencies by Feature Pyramid Network (FPN) and twin transformer, respectively. Meanwhile, compared with the original transformer, the proposed twin transformer is time- and resource-efficient since only a row and a column attention are involved to encode pixels. Moreover, SOTR is easy to be incorporated with various CNN backbones and transformer model variants to make considerable improvements for the segmentation accuracy and training convergence. Extensive experiments show that our SOTR performs well on the MS COCO dataset and surpasses state-of-the-art instance segmentation approaches. We hope our simple but strong framework could serve as a preferment baseline for instance-level recognition. Our code is available at \href{https://github.com/easton-cau/SOTR}{https://github.com/easton-cau/SOTR}.

\end{abstract}

%%%%%%%%% BODY TEXT
\section{Introduction}
Instance segmentation, a fundamental task in computer vision, requires the correct prediction of each object instance and its per-pixel segmentation mask in an image. It becomes more challenging because of the contiguously increasing demands for precise separation of instances in complicated scenes with dense objects and accurate prediction of their masks at the pixel level. Modern instance segmentation approaches \cite{centermask,matterport_maskrcnn_2017} are typically built on CNN and follow the detect-then-segment paradigm, which consists of a detector used to identify and locate all objects, and a mask branch to generate segmentation masks. The success of this segmentation philosophy is attributed to the following favorable merits, i.e. translation equivariance and location, but faces the following obstacles: 1) CNN relatively lacks features' coherence in high-level visual semantic information to associate instances due to the limited receptive field, leading to the sub-optimal results on large objects; 2) Both the segmentation quality and inference speed rely heavily on the object detector, incurring inferior performance in complex scenarios.

\begin{figure}[t]
\begin{center}
\includegraphics[width=1\linewidth]{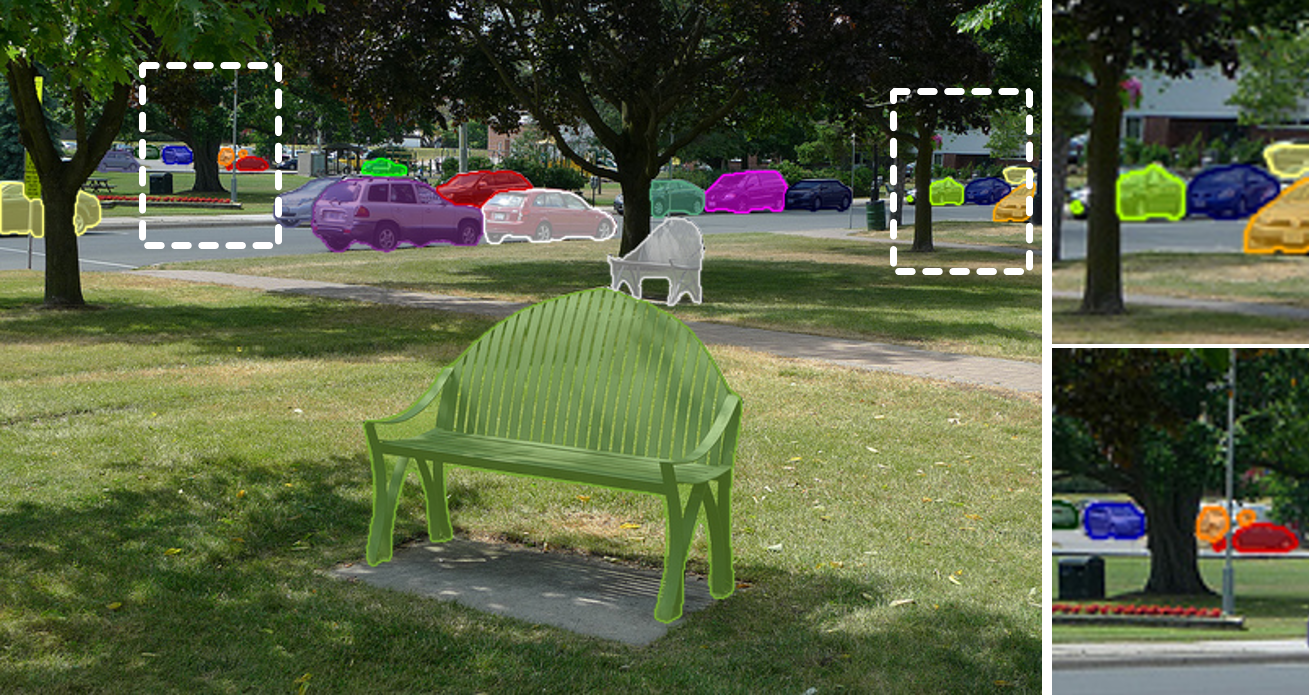}
\end{center}
\vspace{-0.5cm}
   \caption{\textbf{Selected output of SOTR.} We combine CNN with transformer and obtain competitive qualitative results. Notice that not only are larger objects well delineated, targets with elaborate shapes can also get nice segmentation.}
\label{fig1}
\vspace{-1em}
\end{figure}

\begin{figure*}
\begin{center}
%\fbox{\rule{0pt}{2in} \rule{.9\linewidth}{0pt}}
\includegraphics[width=1\linewidth]{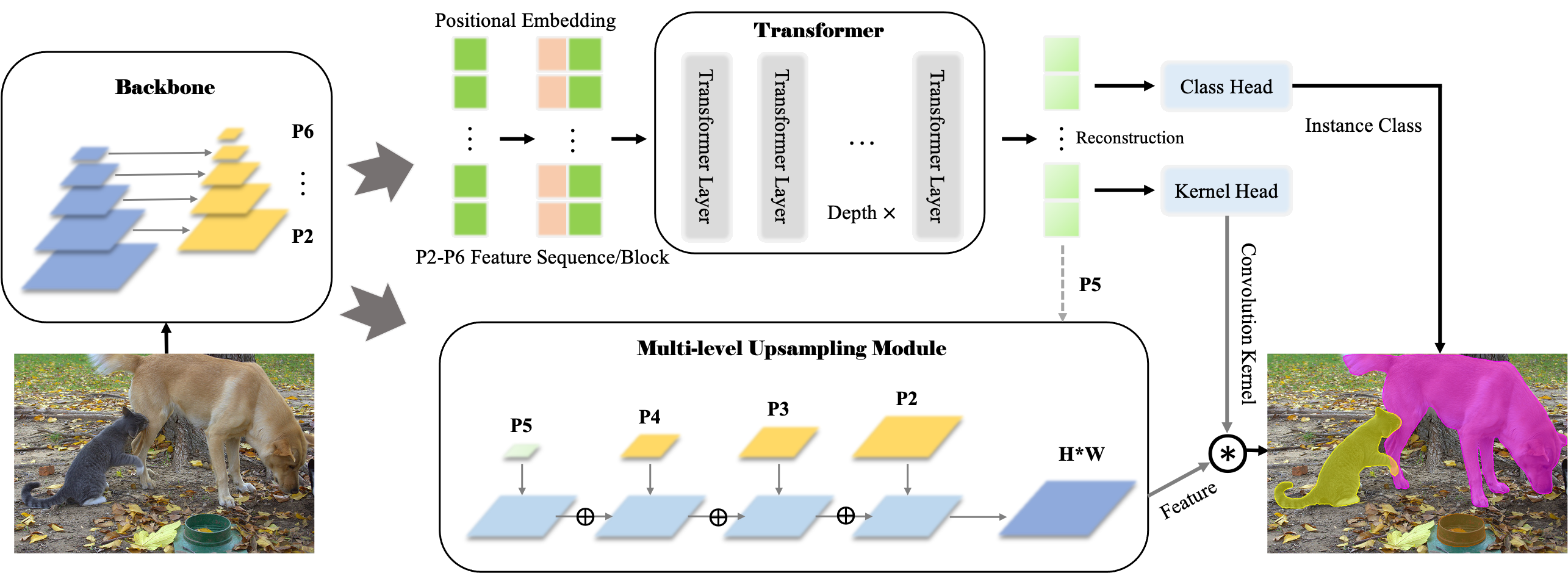}
\end{center}
\vspace{-0.5cm}
   \caption{\textbf{Model overview.} SOTR builds on the simple FPN backbone with minimal modiﬁcation. The model ﬂattens FPN features P2-P6 and supplements them with positional embeddings before feeding them into the transformer model. Two heads are added after the transformer model to predict instance classes and produce dynamic convolution kernels. Multi-level upsampling module take P2-P4 features from FPN and P5 feature from transformer as inputs to generate final masks with dynamic convolution operation by $\circledast$.}
\label{fig2}
\vspace{-1em}
\end{figure*}

To overcome these drawbacks, many recent studies tend to escape from the detect-then-segment manner toward bottom-up strategy \cite{Liu2018AffinityDA,Neven2019InstanceSB}, which learns per-pixel embedding and instance-aware features, and then uses post-processing techniques to successively group them into instances based on the embedding characteristics. Therefore, these methods can well retain position and local-coherence information. However, the main shortcomings of bottom-up models are unstable clustering (e.g., fragmented and joint masks) and poor generalization ability on the dataset with different scenes. Our SOTR (Figure \ref{fig1} and \ref{fig2}) effectively learns position-sensitive features and dynamically generates instance masks following the basic principle of \cite{wang2020solov2}, without the post-processing grouping and the bound of bounding box's locations and scales.

Furthermore, inspired by the power of transformer in natural language processing (NLP) \cite{NEURIPS2020_1457c0d6,devlin-etal-2019-bert,NIPS2017_3f5ee243}, dozens of works try to entirely substitute the convolution operation or combine the CNN-like architectures with transformers for feature extraction in vision tasks \cite{dosovitskiy2020image,chen2020generative,10.1007/978-3-030-58452-8_13}, which can easily capture global-range characteristics and naturally models long-distance semantic dependencies. In particular, self-attention, the key mechanism of transformers, broadly aggregates both feature and positional information from the whole input domain. Thus transformer-based models can better distinguish overlapped instances with the same semantic category, which makes them more suitable than CNN on high-level vision tasks. Nevertheless, insufficiencies still exist in these transformer-based approaches. On the one hand, the typical transformer does not behave well in extracting low-level features, leading to erroneous predictions on small objects. On the other hand, due to the extensive feature map, a large amount of memory and time are required, especially during the training stage. 

To cope with these weaknesses, we propose an innovative bottom-up model called SOTR that ingeniously combines the advantages of CNN and transformer. More specifically, we adopt a new transformer model inspired by \cite{huang2019ccnet} to acquire global dependencies and extract high-level features for predictions in subsequent functional heads. Figure \ref{fig2} shows the overall pipeline of our SOTR. It is composed of three parts, a CNN backbone, a transformer, and a multi-level upsampling module. An image is first fed to FPN to generate feature maps in multi-scale. After patch recombination and positional embedding, the transformer takes the clip-level feature sequences or blocks as inputs and further grasps the global-level semantic features as the powerful complement of the backbone. Then, part of the output feature is input to functional heads for the category and convolution kernel prediction. Finally, the multi-level upsampling module fuses the multi-scale features to a unified one to generate instance masks with the assistance of the dynamic convolution operation.

The focus of SOTR is to investigate ways to better utilize the semantic information extracted by the transformer. With the aim to reduce the memory and computational complexity of the conventional self-attention mechanism, we put forward twin attention, which adopts a sparse representation of the traditional attention matrix. We carry out a great deal of ablation experiments to explore the optimal architecture and hyper-parameters. In summary, not only does our SOTR provide a new framework for instance segmentation, but also it outperforms most of the CNN approaches on the MS COCO \cite{Lin2014MicrosoftCC} dataset. Specifically, the overall contributions of our work are listed as follows:
\begin{itemize}
\item[$\bullet$] We introduce an innovative CNN-transformer-hybrid instance segmentation framework, termed SOTR. It can effectively model local connectivity and long-range dependencies leveraging CNN backbone and transformer encoder in the input domain to make them highly expressive. What's more, SOTR considerably streamlines the overall pipeline by directly segmenting object instances without relying on box detection.
\vspace{-15pt}
\item[$\bullet$] We devise the twin attention, a new position-sensitive self-attention mechanism, which is tailored for our transformer. This well-designed architecture enjoys a significant saving in computation and memory compared with original transformer, especially on large inputs for a dense prediction like instance segmentation.
\vspace{-15pt}
\item[$\bullet$] Apart from pure transformer based models, the proposed SOTR does not need to be pre-trained on large datasets to generalize inductive biases well. Thus, SOTR is easier applied to insufficient amounts of data.
\vspace{-15pt}
\item[$\bullet$] The performance of SOTR achieves $40.2\%$ of AP with the ResNet-101-FPN backbone on the MS COCO benchmark, outperforming most of state-of-the-art approaches in accuracy. Furthermore, SOTR demonstrates significantly better performance on medium ($59.0\%$) and large objects ($73.0\%$), thanks to the extraction of global information by twin transformer.
\end{itemize}

\section{Related work}
\subsection{Instance segmentation}
\textbf{Top-down instance segmentation.} This approach deals with the problem by segmenting after detecting, which takes the inspiration of object detection tasks \cite{R-CNN}. As the representative of anchor-based and two-stage methods, Mask R-CNN \cite{matterport_maskrcnn_2017} added an extra branch for the instance segmentation in the proposed potential bounding boxes on the base of Faster R-CNN \cite{R-CNN}. Also as one of the anchor-based methods, YOLACT \cite{yolact} segmented instances in one-stage but two parallel subtasks: generating prototype masks and predicting mask coefficients for each instance. The final instance mask is the linear combination of the two. On the other hand, some works devoted to generating the segmentation masks within the anchor-free framework. Many of them derived from FCOS \cite{fcos}. For example, CenterMask \cite{centermask} added a novel spacial attention-guided mask branch to FCOS to predict each detected box's segmentation mask.

\textbf{Bottom-up instance segmentation.} Different from top-down segmentation, this approach generates masks by clustering the pixels into each instance in an image. The typical methods include SSAP, SGN, etc. SGN \cite{sgn} solved the problem by decomposing the clustering of instances with three sub-networks.  In addition, the latest bottom-up method SOLO \cite{wang2020solo, wang2020solov2} segmented the instance more directly. Instead of exploiting the relations between pixel pairwise, SOLO handled the clustering problem with classification. It categoried each grid and predicted the mask for each grid end-to-end without clustering. When the scenes are very complicated and dense objects exist in one image, lots of the computation and time will inevitably be lost on background pixels. However, our proposed SOTR takes an image as input, combines CNN with transformer module to extract the features, and directly makes predictions for class probabilities and instance masks.

\subsection{Transformer in vision}
Inspired by the great success of transformer in NLP, researchers propose to apply transformers to solve computer vision problems \cite{han2020survey,khan2021transformers,ho2019axial}. Following the standard transformer paradigms, Dosovitskiy et al. \cite{dosovitskiy2020image} presented a pure transformer model called Vision Transformer (ViT), which achieved state-of-the-art results on image classification tasks. To make the architecture of ViT as similar as the original transformer, the input image was reshaped into a sequence of flattened $2D$ patches and mapped to the corresponding embedding vectors with a trainable linear projection and position embeddings. The pure transformer model could be naturally generalized to produce semantic segmentation by adding a FCN-based mask head. In Segmentation Transformer (SETR) \cite{zheng2020rethinking}, the framework built upon ViT with minimal modification and applied a progressive upsampling strategy as the decoder to generate final masks.

While the above results are encouraging, transformer meets difficulties in extracting low-level features and lacks some inductive biases, so pure transformer models are overly dependent on pre-training on large datasets. These problems can be effectively solved by a combination with CNN backbone. Detection Transformer (DETR) \cite{10.1007/978-3-030-58452-8_13} consisted of a standard CNN backbone and an encoder-decoder transformer for object detection. The former learned a $2D$ representation of the input image and generated the lower-resolution feature map. The latter predicted $N$ objects (box coordinates and class labels) in parallel from the above flatten features with position information. However, two issues exist in DETR. Due to the feature mapping before the relation modeling in transformer, DETR not only suffers from the high computational cost but also performs poorly on small objects. Furthermore, DETR requires a longer training schedule to tune attention weights and focus on meaningful sparse locations.

For instance segmentation, DETR can be extended by appending a mask tower on top of the decoder outputs. Unlike these methods, we rethink the instance segmentation in a different manner and contribute a novel instance segmentation approach assembling CNN and transformer, called SOTR. The differences are apparent. First, SOTR follows standard FCN design and utilizes learnable convolutions to delineate each object region by location, directly segmenting instances in a box-free manner. Second, we adopt twin attention, an alternative self-attention autoregressive block, to signiﬁcantly decrease computation and memory by decomposing the global spatial attention into independent vertical and horizontal attentions.

\begin{figure*}[htb]
\begin{center}
%\fbox{\rule{0pt}{2in} \rule{.9\linewidth}{0pt}}
\includegraphics[width=1.0\linewidth]{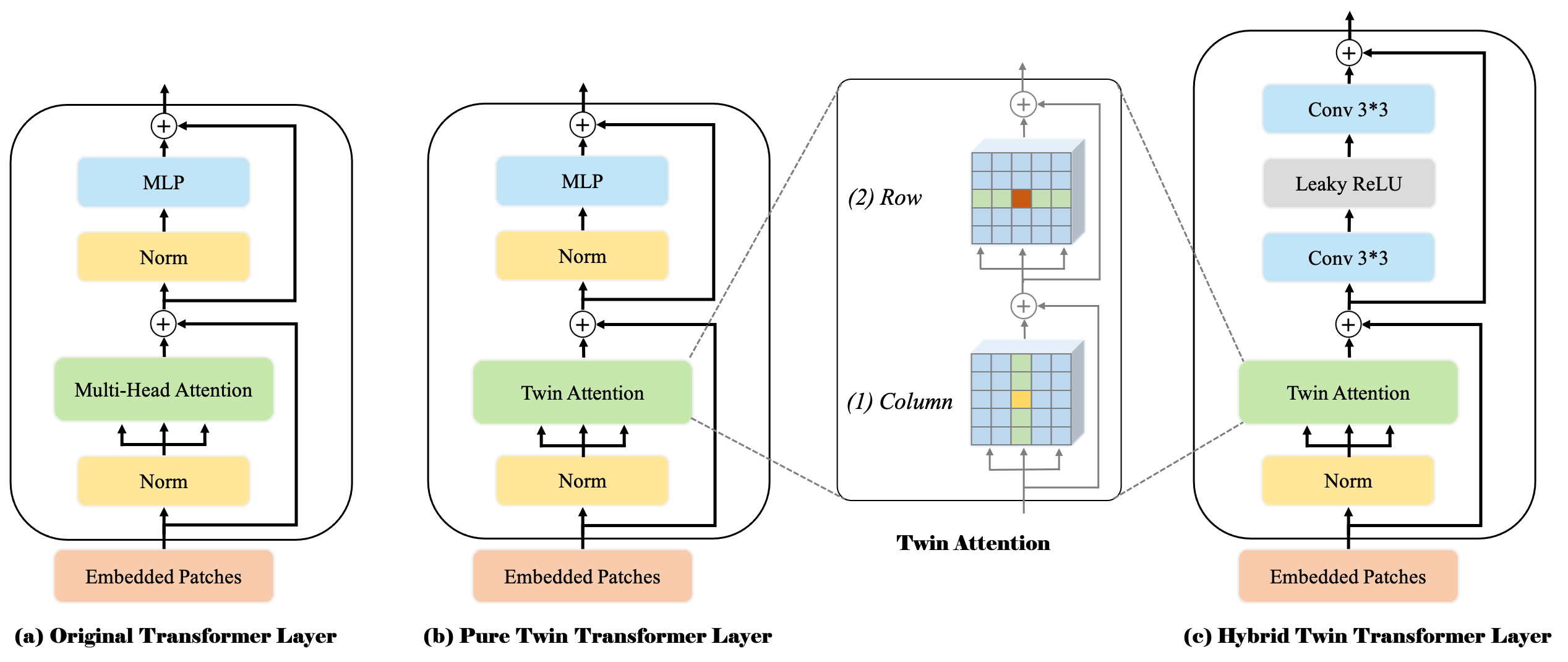}
\end{center}
\vspace{-0.5cm}
   \caption{\textbf{Schematic illustration of three different transformer layer designs.} (a) The original transformer encoder from \cite{NIPS2017_3f5ee243}. To better model long-range dependencies and improve computation efficiency, we introduce different transformer layer designs: (b) pure twin transformer layer and (c) hybrid twin transformer layer. Both layers are based on our designed twin attention that sequentially consists of (1) column- and (2) row-attention.}
\label{fig3}
\vspace{-0.5cm}
\end{figure*}

\section{Methods}
SOTR is a CNN-transformer hybrid instance segmentation model, which can simultaneously learn $2D$ representations and capture long-range information at ease. It follows a direct-segment paradigm, first dividing input feature maps into patches and then predicting per-patch class while dynamically segmenting each instance. Concretely, our model mainly consists of three parts: 1) a backbone to extract image features, especially lower-level and local features, from the input image, 2) a transformer to model global and semantic dependencies, which is appended with functional heads to predict per-patch class and convolution kernel respectively, and 3) a multi-level upsampling module to generate the final segmentation mask by performing dynamic convolution operation between the generating feature map and the corresponding convolution kernel. The overall framework is depicted in Figure \ref{fig2}.

\subsection{Transformer}
\textbf{Twin attention.}
The self-attention mechanism is a key component of transformer models, which inherently captures full-image contexts and learns long-distance interactions between each element in the input sequence. However, self-attention has both quadratic time and memory complicity, incurring higher computational costs on high-dimensional sequences such as images and hindering model scalability in different settings.

To mitigate the above-mentioned problems, we propose the twin attention mechanism to simplify the attention matrix with a sparse representation. Our strategy mainly limits the receptive ﬁeld to a designed block pattern of ﬁxed strides. It first computes the attention within each column while keeping elements in different columns independent. This strategy can aggregate contextual information between the elements on a horizontal scale (see Figure \ref{fig3} (1)). Then, similar attention is performed within each row to fully exploit feature interactions across a vertical scale (shown in Figure \ref{fig3} (2)). The attention in the two scales is sequentially connected to be the final one, which has a global receptive field and covers the information along the two dimensions.

Given feature maps $F_{i} \in \mathbb{R}^{H \times W \times C}$ at layer $i$ of FPN, SOTR first splits the feature maps into $N*N$ patches $P_{i} \in \mathbb{R}^{N \times N\times C}$ and then stacks them into ﬁxed blocks along with the vertical and horizontal directions. Position embeddings are added to the blocks to retain positional information, meaning that the position embedding spaces for the column and row are $1*N*C$ and $N*1*C$. Both of the attention layers adopt the multi-head attention mechanism. To facilitate multi-layer connection and post-processing, all sub-layers in the twin attention produce $N \times N \times C$ outputs. Twin attention mechanism can effectively reduce the memory and computational complexity from standard $\mathcal{O}((H\times W)^2)$ to $\mathcal{O}(H\times W^2+W\times H^2)$\footnote {The memory and computational complexity here is expressed in a more general form of H and W instead of N since the twin attention can process the input of any resolution, not limited to the square tensor.}.

\textbf{Transformer layer.}
In this section, we introduce three different transformer layers based on encoder as our basic building block (as illustrated in Figure \ref{fig3}). The original transformer layer resembles the encoder used in NLP \cite{NIPS2017_3f5ee243} (Figure \ref{fig3} (a)), which comprises two parts: 1) a multi-head self-attention mechanism after a layer normalization \cite{ba2016layer}, and 2) a multi-layer perception after a layer normalization. In addition, a residual connection \cite{he2016identity} is employed to connect these two parts. Finally, a multi-dimensional sequence feature can be obtained as the output of the $K$ serial connection of such transformer layers for subsequent predictions in different functional heads.

In order to make an optimal trade-off between the computational cost and the feature extracting effect, we follow the original transformer layer design and only substitute the multi-head attention with twin attention in the pure twin transformer layer (see Figure \ref{fig3} (b)). In order to further boost the performance of twin transformer, we also design the hybrid twin transformer layer shown in Figure \ref{fig3} (c). It adds two $3\times 3$ convolution layers connected by a Leaky ReLU layer to each twin attention module. It is supposed the added convolution operation can be a useful complement to the attention mechanism, better capturing the local information and enhancing the feature representation.

\textbf{Functional heads.}
The feature maps from transformer modules are input to different functional heads to make subsequent predictions. The class head includes a single linear layer to output a $N\times N\times M$ classification result, where $M$ is the number of classes. Since each patch only assigns one category like YOLO \cite{redmon2016you} for one individual object whose center falls into the patch, we utilize multi-level prediction and share the heads across different feature levels to further improve the model performance and efficiency on objects of different scales. The kernel head is also composed of a linear layer, in parallel with the class head to output a $N\times N\times D$ tensor for subsequent mask generation, where the tensor denotes the $N\times N$ convolution kernels with $D$ parameters. During training, Focal Loss \cite{lin2017focal} is applied to classification while all supervision for these convolution kernels comes from the final mask loss.

\subsection{Mask}
To construct mask feature representations for instance-aware and position-sensitive segmentation, a straightforward way is to make predictions on each feature map of different scales (\cite{wang2020solo, farhadi2018yolov3} and etc.). However, it will increase time and resources. Inspired by the Panoptic FPN \cite{kirillov2019panoptic}, we design the multi-level upsampling module to merge the features from each FPN level and transformer to a unified mask feature. First, the relative low-resolution feature maps P5 with positional information are obtained from the transformer module and combined with P2-P4 in FPN to execute the fusion. For feature maps in each scale, several stages of $3\times 3$ Conv, Group Norm \cite{wu2018group} and ReLU are operated. Then P3-P5 are bilinear upsampled $2\times$, $4\times$, $8\times$, respectively to ($\frac{H}{4}$,$\frac{W}{4}$) resolution. Finally, after the processed P2-P5 are added together, a point-wise convolution and upsampling are executed to create final unified $H \times W$ feature maps.

For instance mask prediction, SOTR generates the mask for each patch by performing dynamic convolution operation on the above unified feature maps. Given predicted convolution kernels $K\in\mathbb{R}^{N\times N\times D}$ from the kernel head, each kernel is responsible for the mask generation of the instance in the corresponding patch. The detailed operation can be expressed as follows:
\begin{equation} \label{conv}
\vspace{0em}
    \begin{aligned}
     Z^{H\times W\times N^2} = F^{H\times W\times C} * K^{N\times N\times D}
    \end{aligned}
\vspace{-0.5em}
\end{equation}

Where $*$ indicates the convolution operation, $Z$ is the final generated mask with a dimension of $H\times W\times N^2$. It should be noted that the value of $D$ depends on the shape of the convolution kernel, that is to say, $D$ equals ${\lambda}^2 C$, where $\lambda$ is kernel size. 
The final instance segmentation mask can be produced by Matrix NMS \cite{wang2020solov2} and each mask is supervised independently by the Dice Loss \cite{milletari2016v}.

\section{Experiments}
We conduct experiments on the challenging MS COCO dataset \cite{Lin2014MicrosoftCC}, which contains $123K$ images with $80$-class instance labels. All models are trained on \texttt{train2017} subset and evaluated on \texttt{test-dev} subset. We also report the standard COCO metrics including average precision (AP), AP at IoU $0.5$ ($\mathrm{AP}_{50}$), $0.75$ ($\mathrm{AP}_{75}$) and AP for objects at different sizes $\mathrm{AP}_{S}$, $\mathrm{AP}_{M}$, and $\mathrm{AP}_{L}$. 

\textbf{Implementation details.} We train SOTR with SGD setting the initial learning rate of $0.01$ with constant warm-up of $1k$ iterations and using weight decay of $10^{-4}$ and momentum of $0.9$. For our ablation experiments, we train for $300K$ iterations with a learning rate drop by a factor of $10$ at $210K$ and $250K$, respectively. Unless specified, all models are trained on $4$ $V100$ GPUs of $32G$ RAM (take about 3-4 days), with batch size $8$. The Python language is used for programming and the deep learning frameworks used are \texttt{PyTorch} and \texttt{Detectron2} \cite{wu2019detectron2}.

\subsection{Ablation experiments}
We carry out a number of ablation experiments on the architectures and hyper-parameters to verify the effectiveness of parameter choice.
 
\begin{table}[htp]
\setlength{\abovecaptionskip}{-0.25cm}
\setlength{\belowcaptionskip}{-0.25cm}
\caption{\textbf{Backbone comparison results.} Better backbone brings expected gains: deeper neural network does better.}
\centering
\vspace{0.3em}
\begin{center}
\begin{small}
%\begin{sc}
\begin{tabular}{ccccc}
\toprule[1.2pt]
Backbone    & $AP$  & $AP_S$   & $AP_M$   & $AP_L$\\
\midrule
Res-50-FPN & 37.5  &9.5  &55.7 & 70.8 \\
Res-101-FPN& \textbf{40.2}  & \textbf{10.3} &\textbf{59.0} & \textbf{73.0} \\
\bottomrule[1.2pt]
\end{tabular}
%\end{sc}
\end{small}
\end{center}
\vskip -0.1in
\label{table:backbone}
\vspace{0em}
\end{table}

\textbf{Backbone architecture.} We compare the performance of the different backbones on extracting features as shown in Table \ref{table:backbone}. We surprisingly find that SOTR with Res-50-FPN can already get $37.5\%$ AP on the COCO and $70.8\%$ AP on large objects. We note that our SOTR automatically benefits from deeper or advanced CNN backbones. In this ablation, results also show that the performance could be further improved by using a better backbone.

\begin{table}[htp]
\setlength{\abovecaptionskip}{-0.25cm}
\setlength{\belowcaptionskip}{-0.25cm}
\caption{\textbf{SOTR incorporating different transformers on COCO \texttt{test-dev}.} Note that all models are trained with same manners including $30K$ iterations, $8$ batch size, etc. Under the same ResNet-101-FPN backbone, the hybrid transformer outperforms all other counterparts.}
\centering
%\vskip 0.15in
\vspace{0.5em}
\begin{center}
\begin{small}
%\begin{sc}
\begin{tabular}{ccccc}
\toprule[1.2pt]
Transformer    & $AP$ & $AP_S$ & $AP_M$ & $AP_L$\\
\midrule
Original     & 37.1 & 9.0  & 56.1 & 71.0 \\
Pure Twin    & 39.7 & 9.9  & \textbf{59.1} & \textbf{73.6}  \\
Hybrid Twin  & \textbf{40.2} & \textbf{10.3} & 59.0 & 73.0 \\
\bottomrule[1.2pt]
\end{tabular}
%\end{sc}
\end{small}
\end{center}
\vskip -0.1in
\label{table:transformer_layer_comparison}
\vspace{0em}
\end{table}

\textbf{Transformer for feature encoding.} We measure the performances of our model with three different transformers. Results of these variants are shown in Table \ref{table:transformer_layer_comparison}. Our proposed pure and hybrid twin transformers surpass the original transformer by a large margin in all metrics, meaning that twin transformer architecture not only successfully captures long-range dependencies across vertical and horizontal dimensions but also is more suitable to be combined with CNN backbone to learn features and representations for images. For pure and twin transformers, the latter works much better. We assume the reason is that $3*3$ Conv can extract the local information and improve feature expression to enhance the rationality of the twin transformer.

\begin{table}[htp]
\setlength{\abovecaptionskip}{-0.25cm}
\setlength{\belowcaptionskip}{-0.25cm}
\caption{\textbf{Comparison of different depth.} Performance with Res-101-FPN backbone. Original and Twin denote the original transformer and the hybrid twin transformer.}
\centering
%\vskip 0.15in
\vspace{0.5em}
\begin{center}
\begin{small}
%\begin{sc}
\begin{tabular}{ccccc}
\toprule[1.2pt]
Transformer                 & Depth & AP & Time(ms) & Memory\\
\midrule
\multirow{2}*{Original}     & 6 & 36.2  & 147 & 6907M \\
                            & 12& 37.1  & 199 & 10696M\\
\midrule
\multirow{2}*{\textbf{Twin}}& 6 & 37.6  &\textbf{113} &\textbf{3778M}  \\
                            & \textbf{12}& \textbf{40.2} & 161 & 5492M\\
\bottomrule[1.2pt]
\end{tabular}
%\end{sc}
\end{small}
\end{center}
\vskip -0.1in
\label{table:transformer_comparison}
\vspace{0em}
\end{table}

\textbf{Transformer depth.} To verify the effect of transformer depth on SOTR, we conduct ablation experiments on the original transformer and the hybrid twin transformer, respectively. As shown in Table \ref{table:transformer_comparison}, both sets of experiments show that by increasing the depth of the transformer, we can increase AP while sacrificing inference time and memory. Also, the twin transformer brings $3.1\%$ AP gains in contrast to the original transformer and reduces the memory footprint by about $50\%$, which shows the superiority of this structure. However, as the transformer goes deeper, the attention maps gradually become similar, i.e., attention collapse. So transformer fails to learn more effective features and hinders model from getting expected performance gain. In further ablation experiment, we use hybrid twin transformer with \texttt{depth=12} for our baseline model if not specified.

\begin{table}[htp]
\setlength{\abovecaptionskip}{-0.25cm}
\setlength{\belowcaptionskip}{-0.25cm}
\caption{\textbf{Feature map substitution on multi-level upsampling process.} The check mark indicates whether to substitute P4 or P5 layer with feature maps generated by the 12-layer hybrid twin transformer.}
\centering
%\vskip 0.15in
\vspace{0.5em}
\begin{center}
\begin{small}
%\begin{sc}
\begin{tabular}{cccccc}
\toprule[1.2pt]
P4  &P5     & $AP$ & $AP_S$ & $AP_M$ & $AP_L$\\
\midrule
   &          & 38.8 & 9.7  & 58.0 & 72.0 \\
   &\checkmark     &  \textbf{40.2} & \textbf{10.3} & 59.0   & 73.0 \\
 \checkmark&\checkmark &39.9 & 10.1 & \textbf{59.1} & \textbf{73.7} \\
\bottomrule[1.2pt]
\end{tabular}
%\end{sc}
\end{small}
\end{center}
\vskip -0.1in
\label{table:positional_embedding}
\vspace{0em}
\end{table}

\begin{figure}[htp]
\vspace{0.5em}
\setlength{\belowcaptionskip}{-0.5cm}
\begin{center}
\includegraphics[width=1\linewidth]{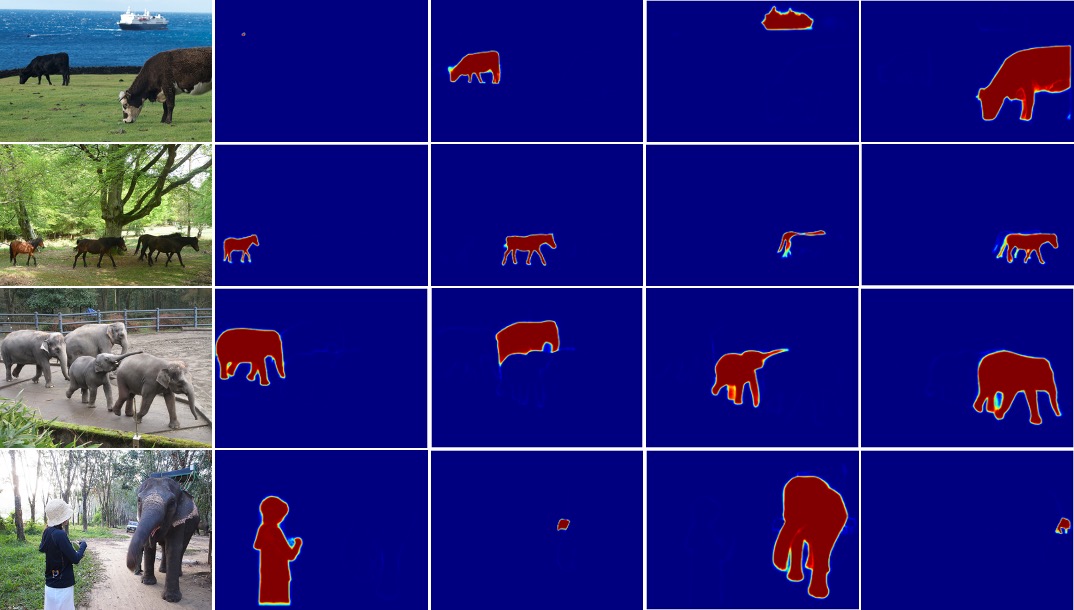}
\end{center}
\vspace{-1em}
   \caption{\textbf{SOTR behavior.} We show the visualization of the mask feature. For each row, the left side is the original picture, and the right side is the positional-sensitive mask correspond to it.}
\label{figextra}
\vspace{1em}
\end{figure}

\textbf{Multi-level Upsampling Module.} In this ablation experiment, we explore the effect of feature maps generated by transformer on the multi-level upsampling module. As shown in Table \ref{table:positional_embedding}, the model has the highest AP value when only replacing the P5 layer from FPN with the feature map generated by transformer. When replacing both the P4 and P5 layers, the model's AP value has a slight drop ($-0.3\%$). This shows that using generated feature maps on more layers does not bring noticeable improvement in overall AP, and the P5 from transformer already enables predictions to be well position-sensitive. The reason why $AP_M$ and $AP_L$ are slightly improved 0.1\% and 0.7\% is that the P4 from transformer carries more global and larger object features than the P4 from FPN. In addition, SOLOv2 employs Coordconv on P5 to add positional information for segmentation, while SOTR substitutes it with transformers to gain such information and generate position-sensitive feature maps as shown in Figure \ref{figextra}.

\begin{table}[htp]
\setlength{\abovecaptionskip}{-0.25cm}
\setlength{\belowcaptionskip}{-0.25cm}
\caption{\textbf{Dynamic convolution kernel vs. Static convolution kernel.} Learnable convolution kernel can considerably improve the results.}
\centering
%\vskip 0.15in
\vspace{0.5em}
\begin{center}
\begin{small}
%\begin{sc}
\begin{tabular}{cccccc}
\toprule[1.2pt]
Twin & DCK      & $AP$ & $AP_S$ & $AP_M$ & $AP_L$\\
\midrule
\checkmark   &     & 38.6 & 9.5  & 57.7 & 72 \\
& \checkmark     & 39.7 & \textbf{17.3} & 42.9 & 57.4 \\
\checkmark& \checkmark     &  \textbf{40.2} & 10.3 & \textbf{59.0} & \textbf{73.0} \\
\bottomrule[1.2pt]
\end{tabular}
%\end{sc}
\end{small}
\end{center}
\vskip -0.1in
\label{table:dynamic_convolution}
\vspace{-2em}
\end{table}

\begin{figure*}
\begin{center}
%\fbox{\rule{0pt}{2in} \rule{.9\linewidth}{0pt}}
\includegraphics[width=\linewidth]{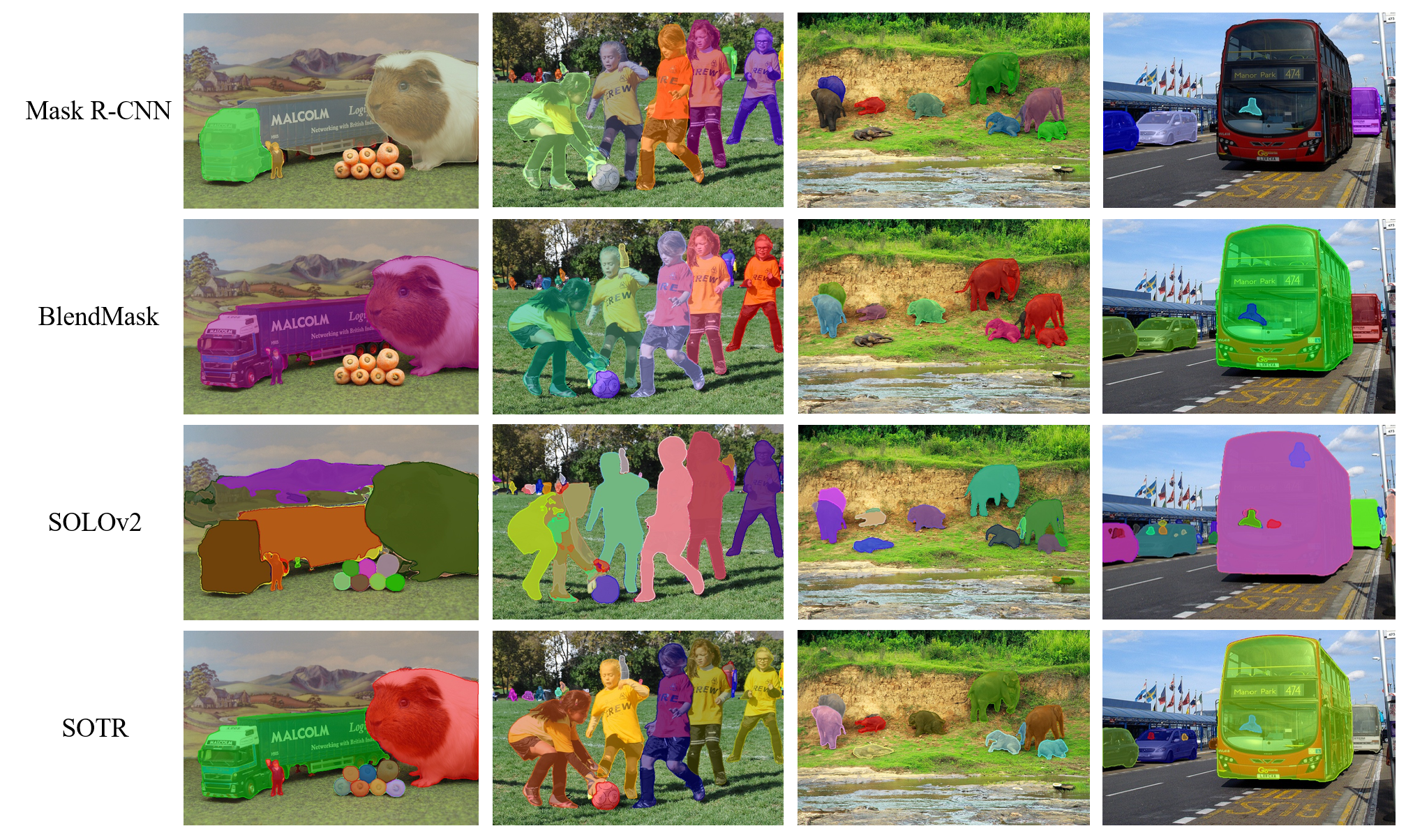}
\end{center}
\vspace{-0.5cm}
   \caption{\textbf{Detailed comparison with other methods.} We compare the segmentation result of our method against Mask R-CNN \cite{matterport_maskrcnn_2017}, Blendmask \cite{blendmask} and SOLOv2 \cite{wang2020solov2}.The code and the trained model are provided by the original author. All models use ResNet-101-FPN as the backbone and are based on \texttt{Pytorch} and \texttt{Detectron2}. Our masks are of typically higher quality.}
\label{fig5}
\vspace{-0.5em}
\end{figure*}

\textbf{Dynamic convolution.} 
For mask generation, we have two options: to directly output instance masks in the static convolutional manner or to continually segment objects by dynamic convolution operation. The former one does not require the extra functional head to predict convolution kernels, while the latter one includes convolution kernel to generate final masks with the assistance of the fused feature. We compare these two modes in Table \ref{table:dynamic_convolution}. 
\begin{table}[htp]
\setlength{\abovecaptionskip}{-0.25cm}
\setlength{\belowcaptionskip}{-0.25cm}
\caption{\textbf{Real-time setting comparison.} Metrics for models are obtained using their official trained models. The speed is reported on a single V100 GPU on COCO.}
\centering
%\vskip 0.15in
\vspace{0.5em}
\begin{center}
\begin{small}
%\begin{sc}
\setlength{\tabcolsep}{1mm}{
\begin{tabular}{cccccc}
\toprule[1.2pt]
Model         & Backbone      & \#param.  & FLOPs    & FPS    & AP   \\
\midrule
YOLACT-550    & R-50-FPN      &  140.23M  & 61.59G   & \textbf{44.1}   & 28.2 \\
PolarMask-600 & R-50-FPN      &   \textbf{34.46M}  & 81.83G   & 21.7   & 27.6 \\
SOTR-RT-736   & R-50-FPN      &   38.20M  & \textbf{60.31G}   & 25.2   & \textbf{30.7} \\
\bottomrule[1.2pt]
\end{tabular}}
%\end{sc}
\end{small}
\end{center}
\vskip -0.1in
\label{table:6}
% \vspace{-1.5em}
\vspace{-1.8em}
\end{table}
As shown, SOTR without twin transformer achieves an $AP$ of $39.7\%$, indicating the twin transformer brings $0.5\%$ gains. In addition, the dynamic convolution strategy could improve the performance by almost $1.5\%$ AP. We explain as follows. On the one hand, dynamic convolution significantly boosts the representation capability due to the non-linearity. On the other hand, dynamic convolution contribute to better and faster convergence in training than its static counterpart. 

% As illustrated in Figure \ref{fig4}, the loss curve of static convolution oscillates more severely and converges at a higher loss value, while the loss of dynamic convolution is more stable and converges at about $0.4\%$.

\textbf{Real-time model and comparison.}
SOTR focuses on boosting accuracy of instance segmentation while can be modified to be a real-time (RT) model with some sacrifice of the accuracy. The number of transformer layers of our designed SOTR-RT is reduced to two and the input shorter side is 736. Table \ref{table:6} shows the performance of SOTR-RT models compared with others.

\begin{table*}[htp]
\setlength{\belowcaptionskip}{-0.5cm}
\caption{\textbf{Quantitative results.} mask AP(\%) on COCO \texttt{test-dev}. We compare our SOTR with state-of-the-art instance segmentation methods, and all entries in the table are \textsl{single-model} results. We denote the backbone architecture with \texttt{network-depth-feature}, where \texttt{Res} refer to ResNet \cite{he2016deep}. Mask R-CNN* is the improved version with scale augmentation and longer training time($6\times$). }
\label{table:instance_segmentation}
\vskip 0.15in
\vspace{-0.7cm}
\begin{center}
\begin{small}
%\begin{sc}
\begin{tabular}{cccccccc}
\toprule[1.2pt]
Method & Backbone        & $AP$ & $AP_{50}$ & $AP_{75}$  & $AP_S$   & $AP_M$ &  $AP_L$ \\
\midrule
FCIS \cite{FCIS}       & Res-101-C5    &29.5 &51.5 &30.2 &8.0  &31.0 &49.7\\
MaskLab+ \cite{Chen2018MaskLabIS}   & Res-101-C4    &37.3 &59.8 &39.6 &16.9 &39.9 &53.5\\
Mask R-CNN \cite{matterport_maskrcnn_2017} & Res-101-FPN   &35.7 &58.0 &37.8 &15.5 &38.1 &52.4 \\
Mask R-CNN*& Res-101-FPN   &37.8 &59.8 &40.7&\textbf{20.5} &40.4 &49.3\\
RetinaMask \cite{Fu2019RetinaMaskLT} & Res-101-FPN   &34.7 &55.4 &36.9 &14.3 &36.7 &50.5\\
MS R-CNN \cite{MS-RCNN}   & Res-101-FPN   &38.3 &58.8 &41.5 &17.8 &40.4 &54.4\\
TensorMask \cite{Chen2019TensorMaskAF} & Res-101-FPN   &37.1 &59.3 &39.4 &17.4 &39.1 &51.6 \\
ShapeMask \cite{Kuo2019ShapeMaskLT}&Res-101-FPN &37.4&58.1&40.0&16.1&40.1&53.8\\
YOLACT \cite{yolact}     & Res-101-FPN   &31.2 &50.6 &32.8 &12.1 &33.3 &47.1\\
YOLACT++ \cite{yolact++} & Res-101-FPN   &34.6 &53.8 &36.9 &11.9 &36.8 &55.1\\
PolarMask \cite{Xie2020PolarMaskSS}  & Res-101-FPN   &32.1 &53.7 &33.1 &14.7 &33.8 &45.3\\
SOLO \cite{wang2020solo}    & Res-101-FPN      &37.8 &59.5 &40.4 &16.4 &40.6 &54.2\\ 
BlendMask \cite{blendmask}   & Res-101-FPN  &38.4 &60.7 &41.3 &18.2 &41.5 &53.3\\
CenterMask \cite{Wang2020CenterMaskSS}  & Hourglass-104&34.5 &56.1 &36.3 &16.3 &37.4 &48.4\\
MEInst \cite{MEInst}     & Res-101-FPN   &33.9 &56.2 &35.4 &19.8 &36.1 &42.3\\
SOLOv2 \cite{wang2020solov2}    & Res-101-FPN    &39.7 &60.7 &42.9 &17.3 &42.9 &57.4\\
\textbf{SOTR}     & Res-101-FPN      &\textbf{40.2} &\textbf{61.2} &\textbf{43.4} &10.3 &\textbf{59.0} &\textbf{73.0}\\
\midrule
SOLOv2 \cite{wang2020solov2}    & Res-DCN-101-FPN    &41.7 &63.2 &45.1 &\textbf{18.0} &45.0 &61.6\\
\textbf{SOTR}     & Res-DCN-101-FPN  &\textbf{42.1} &\textbf{63.3} &\textbf{45.5} &11.5 &\textbf{60.8} &\textbf{74.4}\\
          
\bottomrule[1.2pt]
\end{tabular}
%\end{sc}
\end{small}
\end{center}
\vskip -0.1in
\vspace{-1.5em}
\end{table*}

\subsection{Main result}
\textbf{Quantitative results}
We compare SOTR to the state-of-the-art methods in instance segmentation on MS COCO \texttt{test-dev} in Table \ref{table:instance_segmentation}. SOTR with ResNet-101-FPN achieves a mask AP of $40.2\%$, which is much better than other modern instance segmentation methods. Compared with the traditional two-stage instance segmentation method Mask R-CNN \cite{matterport_maskrcnn_2017}, SOTR has achieved better prediction accuracy ($+2.4\%$ AP), and has achieved a considerable improvement in medium and large targets($+20.9\%$ in $AP_M$ and $+20.6\%$ in $AP_L$). As one of the box-free algorithms, SOTR also has a significant improvement compared to SOLO \cite{wang2020solo} and PolarMask \cite{Xie2020PolarMaskSS}.
Furthermore, to the best of our knowledge, SOTR is the first method to achieves nearly $60\%$ AP in medium objects and over $70\%$ AP in medium and large objects.

\textbf{Qualitative results}
We compare SOTR with the official Mask R-CNN \cite{{matterport_maskrcnn_2017}}, BlendMask \cite{blendmask} and SOLOv2 \cite{wang2020solov2} models with ResNet-101 backbone \footnote{To make a fair comparison with Mask R-CNN, BlendMask, and SOLOv2, the code base we use is \texttt{Detectron2}. Recently released \texttt{Detectron2} originates from \texttt{maskrcnn\_benchmark} with significant enhancements for performance.}. The segmentation masks are displayed in Figure \ref{fig5}. Our SOTR performs better than Mask R-CNN and BlendMask in two cases: 1) objects with elaborate shapes easy to be omitted by other models (e.g. carrots in front of the train, recumbent elephants, drivers in the small cars), Mask R-CNN and BlendMask fail to detect them as positive instances. 2) objects overlapping with each other (e.g. person in front of the train), the two cannot separate them with accurate borders. SOTR is able to predict mask with sharper borders, while SOLOv2 tends to segment targets into separate parts (e.g. dividing the train into the head and the body) and sometimes cannot exclude the background from the image. Due to the introduction of transformer, SOTR can better gain a comprehensive global-information to avoid such split on objects. Furthermore, compared with SOTR, SOLOv2 has a high false positive rate by assigning nonexistent objects as instance.

\section{Conclusion}
In this paper, we have proposed a new direct-segment instance segmentation method built upon CNN and transformer, which dynamically predicts the segmentation mask of each instance without object detectors, simplifying the overall pipeline. In order to process large inputs that are organized as multidimensional tensors, we design a transformer model variant based on the twin attention mechanism, which successfully reduces the memory and computational complexity to $\mathcal{O}(H\times W^2+W\times H^2)$. In addition, it is efﬁcient and easy to be integrated with different mainstream CNN backbones. Extensive ablation studies are conducted to verify the core factors of SOTR. Without bells and whistles, SOTR with ResNet-101-FPN backbone performs well and achieves $40.2\%$ mAP on MS COCO dataset. We believe that our SOTR is capable of serving as a fundamental and strong baseline for instance segmentation tasks.

\section{Acknowledgments} 
The work is supported by National Key R$\&$D Program of China (2020YFD0900204) and Key-Area Research and Development Program of Guangdong Province China (2020B0202010009).

\newpage
{\small
\normalem
\bibliographystyle{ieee_fullname}
\bibliography{egbib}
}

\end{document}